\def\year{2020}\relax
\documentclass[letterpaper]{article} 
\usepackage{aaai20}  
\usepackage{times}  
\usepackage{helvet} 
\usepackage{courier}  
\usepackage[hyphens]{url}  
\usepackage{subcaption}
\usepackage{graphicx} 
\usepackage{graphicx}  
\title{Tree Search vs Optimization Approaches for Map Generation}

\author{Debosmita Bhaumik\textsuperscript{\rm 1}\thanks{Both authors contributed equally to this research.}, Ahmed Khalifa\textsuperscript{\rm 1}\setcounter{footnote}{0}\footnotemark, Michael Cerny Green\textsuperscript{\rm 1,2}, Julian Togelius\textsuperscript{\rm 1,2,3}\\ 
\textsuperscript{\rm 1}New York Univeristy, \textsuperscript{\rm 2}OriGen.AI, \textsuperscript{\rm 3}Modl.AI\\
debosmita.bhaumik01@gmail.com, ahmed@akhalifa.com, mike.green@nyu.edu, julian@togelius.com 
}

\begin{document}

\maketitle

\begin{abstract}
Search-based procedural content generation uses stochastic global optimization algorithms to search for game content. However, standard tree search algorithms can be competitive with evolution on some optimization problems. We investigate the applicability of several tree search methods to level generation and compare them systematically with several optimization algorithms, including evolutionary algorithms. We compare them on three different game level generation problems: Binary, Zelda, and Sokoban. We introduce two new representations that can help tree search algorithms deal with the large branching factor of the generation problem.
We find that in general, optimization algorithms clearly outperform tree search algorithms, but given the right problem representation certain tree search algorithms performs similarly to optimization algorithms, and in one particular problem, we see surprisingly strong results from MCTS.
\end{abstract}

\section{Introduction}
Generating levels for games is a research problem with broad relevance across most game genres and many domains outside of games. Video games, from shooters to role-playing games to puzzle games, need level generation in order to create larger and more replayable games, adapt games to players, simplify game development, and enable certain kinds of aesthetics. Domains such as architecture, urban planning, military simulation and logistics need scenario and environment generation for similar reasons, and these problems are often very similar to game level generation. In reinforcement learning, level generation allows for creating variable environments which helps with generalization~\cite{justesen2018illuminating}. For these reasons, the past decade has seen considerable interest in research on level generation and other forms of procedural content generation (PCG)~\cite{shaker2016procedural}.

One approach to the generation of levels as well as other types of game content is to use evolutionary algorithms or similar global stochastic optimization algorithms to search for good levels. This approach, called search-based PCG, requires that the levels are represented in such a way that the level space can be efficiently searched, and that there is a fitness function which can reliably approximate content quality~\cite{togelius2011search}. 

As an alternative to using evolutionary methods, tree search methods such as Monte Carlo Tree Search (MCTS)~\cite{browne2012survey} have been suggested for PCG. 
While it seems that both stochastic optimization and tree search can be used for level generation (and many related generative tasks), we have only been able to find a handful of papers doing PCG by tree search (see next section). Given the very different ways in which these algorithm types search, it stands to reason that they should differ sharply in performance depending on the objective and representation. Maybe there are domains where tree search significantly outperform optimization methods?

This paper systematically compares several tree search algorithms with optimization algorithms on three different 2D level generation problems, across three different representations. We attempt to answer the question which tree search algorithms can have similar or better performance with optimization for level generation, and when.

\section{Background}\label{sec:background}
Procedural content generation (PCG) is the automatic generation of game content, be it game levels, characters, quests, storylines, game elements like trees and rocks, or even entire games themselves~\cite{shaker2016procedural}. Search-based PCG is a subset of PCG methods that relies on search or optimization methods~\cite{togelius2011search}. In practice, evolutionary algorithms are most commonly used. This section describes previous research in the areas of tree search and evolution as well as procedural content which can be generated using these methods. 

\subsection{Tree Search}
Tree search algorithms try to find solutions by starting at a root node and expanding child nodes in a systematic way.
Popular techniques include Breadth-First Search (BFS), Depth-First Search (DFS), Greedy Best First Search (GBFS), and Monte Carlo Tree Search (MCTS)~\cite{russell2016artificial,browne2012survey}. Tree search agents are commonly used in game-playing agents, like for the Mario AI Benchmark~\cite{togelius20102009}, Chess~\cite{campbell2002deep}, Go~\cite{gelly2006modification,silver2016mastering}, and general video games~\cite{perez2016general} among many others.

In the area of PCG, few examples exist of using tree search to generate game content. \citeauthor{browne2013uct}~\shortcite{browne2013uct} first explored this concept by using a variant of the Upper Confidence Bound for Trees equation (UCT) called \textit{Upper Confidence Bounds for Graphs} (UCG) to develop biominoes, simple polyomino\footnote{Orthogonally connected sets of squares~\cite{golomb1996polyominoes}.} shapes and the Pentominoes puzzle domain. \citeauthor{summerville2015mcmcts}~\shortcite{summerville2015mcmcts} generated levels for Super Mario Bros (Nintendo 1985) using Markov Chains where the exploration was guided using Monte Carlo Tree Search. \citeauthor{kartal2013generating}~\shortcite{kartal2013generating} used MCTS to generate stories, taking advantage of MCTS' ability to successfully navigate the large search spaces associated with possible character actions and reactions within narratives.  \citeauthor{kartal2016generating}~\shortcite{kartal2016generating} also used MCTS to generate \textit{Sokoban} (Imabayashi 1981) levels. At each node in the MCTS tree, the level generator is given choices to take to modify the level, such as deleting/adding objects and moving an agent around within the level to simulate gameplay. \citeauthor{graves2016procedural}~\shortcite{graves2016procedural} experimented with using MCTS to generate Angry Birds (Rovio Entertainment 2009) levels. At each node in the tree, the level generator can place/remove structures/pigs or do nothing at all. Finally, exhaustive search can also be used to create all possible content artifacts in some space~\cite{sturtevant2018exhaustive}.

\subsection{Optimization}
Global optimization algorithms are algorithms which focus only on finding a good solution, which maximizes or minimizes some objective, not on the path leading from an origin state to that solution. Evolutionary algorithms, a family of stochastic population-based algorithms, are a good representative of this class. Such algorithms are popular choices for PCG because it is easy to frame the PCG as a single-point or population-based optimization problem, where the fitness functions/objectives can be cleanly mapped to game elements like difficulty, time, physical space, level variety, etc
~\cite{togelius2011search}. Ashlock did this several ways, such as optimized puzzle generation for different difficulties~\cite{ashlock2010automatic}, or stylized cellular automata evolution for cave generation~\cite{ashlock2015evolvable}. \citeauthor{mcguinness2011decomposing}~\shortcite{mcguinness2011decomposing} created a micro-macro level generation process, using a wide variety of fitness functions based on level elements. \citeauthor{shaker2013evolving}~\shortcite{shaker2013evolving} evolved levels for \emph{Cut the Rope} (ZeptoLab 2010) using constrained evolutionary search where the fitness measures the playability using playable agents. In addition to evolving level elements in GVGAI~\cite{khalifa2016general}, PuzzleScript~\cite{khalifa2015automatic}, and Super Mario Bros~\cite{khalifa2019intentional}, \citeauthor{khalifa2015literature}~\shortcite{khalifa2015literature} offers a literature review of search based level generation within puzzle games. 

\section{Methods}
In this paper, we compare tree search algorithms to optimization algorithms. We decided to compare them over three different problems (Binary, Zelda and Sokoban) which were introduced as part of the PCGRL framework~\cite{khalifa2020pcgrl}. We settled on these problems as they cover different types of games and heuristic/fitness functions already exist. Also, variants of these generation problems had been tackled before using optimization algorithms~\cite{ashlock2011search,ashlock2018exploring,khalifa2016general,charity2020mech,khalifa2015automatic}, meaning that we already know of effective problem representations for optimization.
For all the used algorithms, the tree search algorithm stops either when it finds a solution to the problem or time runs out. In this section, we will talk about the different algorithms, representations, problems used in this work.

\subsection{Tree Search Algorithms}\label{sec:tree-search}
We compare four simple and commonly used tree search algorithms, two of which are uninformed search (Breadth First Search and Depth First Search) and two are informed search (greedy best first search and Monte Carlo Tree Search). For the first three algorithms we use the canonical versions as defined in~\cite{russell2016artificial}.

\textbf{Breadth First Search(BFS)} is a simple uninformed search algorithm that expands a full tree level before exploring deeper nodes using a queuing system.

\textbf{Depth First Search (DFS)} is another uninformed search algorithm that always expands one of the nodes at the deepest level of the tree using a stack system. When the search hits a dead end, it goes back and expands nodes at shallower levels. 

\textbf{Greedy Best First Search (GBFS)} is an informed search algorithm which uses the cost to reach the goal as heuristic function and a priority queue to select the most promising nodes in the search tree first. 

\textbf{Monte Carlo Tree Search (MCTS)}\label{sec:tree-search-mcts} is a stochastic tree-search algorithm~\cite{browne2012survey} that creates asymmetric trees by expanding the most promising branches of the search space using random sampling (rollout). There are many variants of MCTS~\cite{browne2012survey}; we use UCT~\cite{kocsis2006improved}; arguably the most widely used version.
UCT uses the UCB1 equation to balance between exploration and exploitation: 
\[
    UCB1_{i} = \frac{V_{i}}{n_{i}} + c \sqrt{\frac{\ln{N}}{n_{i}}}
\]
where $V_{i}$ is the total accumulated rewards for that node, $n_{i}$ is the total number of visits for that node, $N$ is the total number of visits of the parent node, and $c$ is a constant balancing between the exploitation term (first term) and exploration term (second term).

\subsection{Optimization Algorithms}\label{sec:optimization}
For the optimization algorithms, we explore two single-point optimization algorithms (hill climbing and simulated annealing) and two population-based optimization algorithms (evolution strategy and genetic algorithm).

\textbf{Hill Climbing (HC)} is a single-point optimization algorithm~\cite{russell2016artificial} that initializes a random solution and keeps improving the solution (by comparing it to all the possible neighbors) until a local optimal solution is found. 

\textbf{Simulated Annealing (SA)} is a single-point global optimization algorithm~\cite{russell2016artificial} that tries to find a global optimum in the presence of several local optima. Instead of always accepting a better neighbor, it can accept less optimal neighbors with probability less than 1.
The probability is calculated by $P = exp(-d/T)$, where $d$ is the absolute difference between the current solution's score and the new solution's score and $T$ is temperature. Temperature is initially given a high value which slowly decreases every iteration using a cooling rate $c$ ($T = T * c$).

\textbf{Evolution Strategy (ES)} is a nature inspired population-based optimization algorithm~\cite{brownlee2011algorithms}. It applies selection and mutation operators to a population, that contains solutions, to evolve better and better solutions. The process begins with a random population of $\mu + \lambda$ individuals and calculates the fitness of the entire population using a fitness function. $\lambda$ worst individuals are removed from the population and replaced with mutated version of the top $\mu$ individuals. 

\textbf{Genetic Algorithm (GA)} is a population-based optimization technique inspired by the Darwinian principle of evolution~\cite{brownlee2011algorithms}. Like ES, it uses nature inspired operators like mutation and selection as well as a crossover operation to generate high quality solutions. Starting with a random population, it selects individuals based on their fitness for reproduction. These individuals produce a new solution using crossover and mutation operators (with different probabilities) which is inserted into the next population. Additionally, the most fit individuals are immediately inserted into the next generation in a process known as \emph{elitism}. The reproductive process goes on until the new population is fully created.


\begin{figure*}
    \centering
    \begin{subfigure}[t]{0.25\linewidth}
        \centering
        \includegraphics[width=\linewidth]{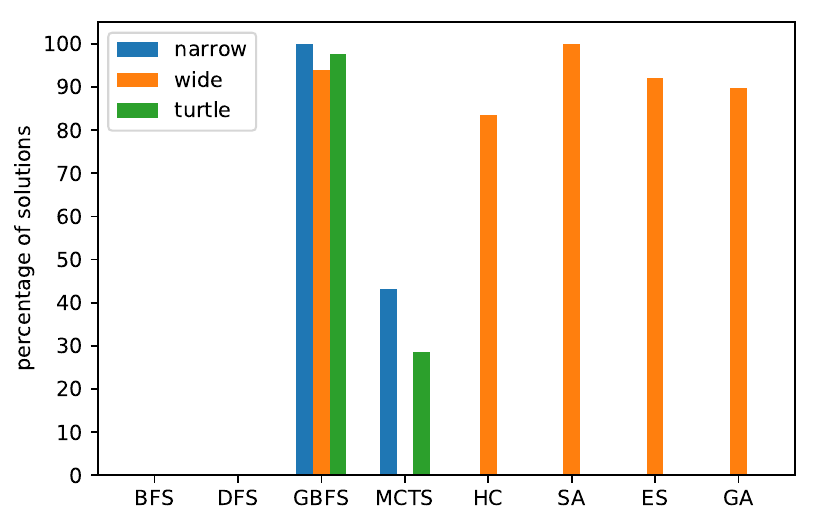}
        \caption{Binary}
        \label{fig:binary_solution}
    \end{subfigure}
    \begin{subfigure}[t]{0.25\linewidth}
        \centering
        \includegraphics[width=\linewidth]{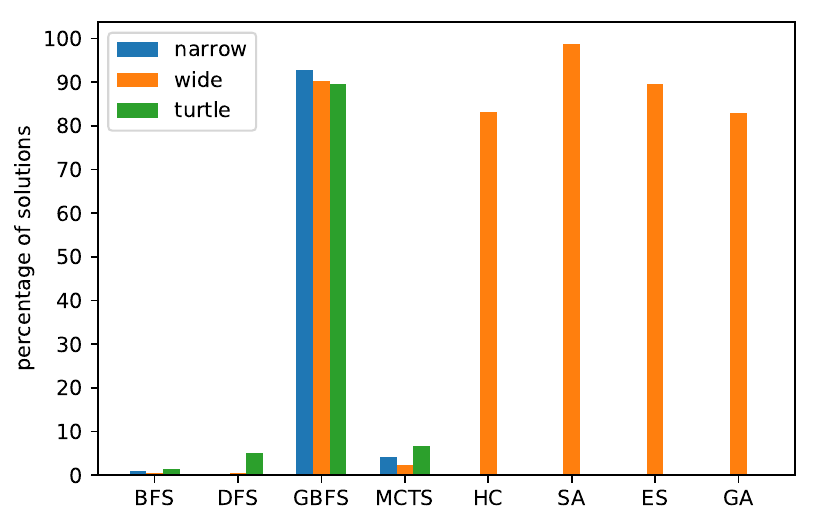}
        \caption{Zelda}
        \label{fig:zelda_solution}
    \end{subfigure}
    \begin{subfigure}[t]{0.25\linewidth}
        \centering
        \includegraphics[width=\linewidth]{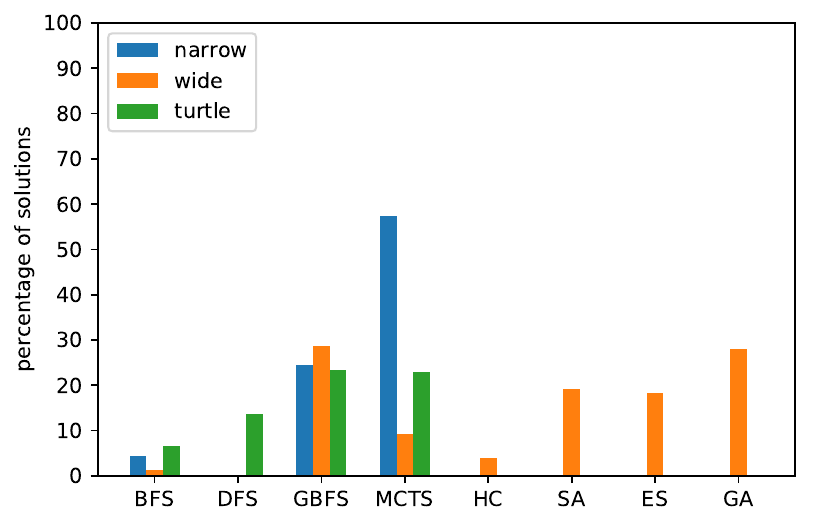}
        \caption{Sokoban}
        \label{fig:sokoban_solution}
    \end{subfigure}
    \caption{The playability percentage for all the algorithms over Binary, Zelda, and Sokoban problems.}
    \label{fig:solutions}
\end{figure*}

\begin{figure*}
    \centering
    \begin{subfigure}[t]{0.25\linewidth}
        \centering
        \includegraphics[width=\linewidth]{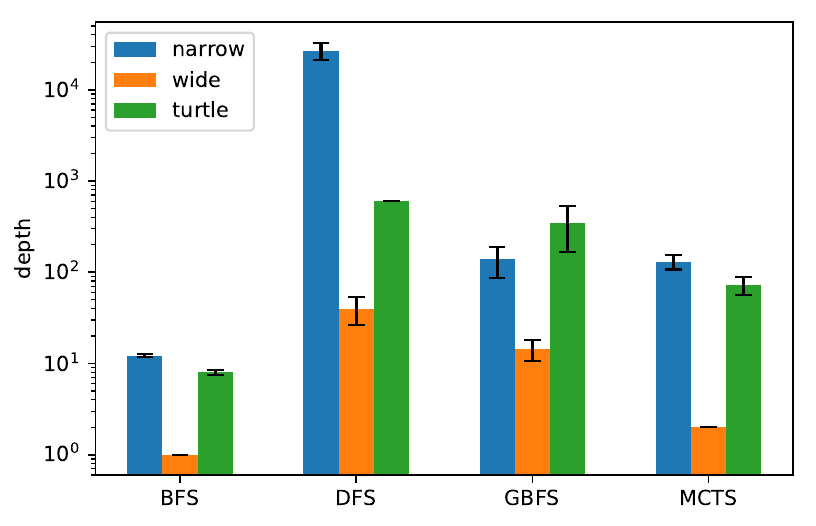}
        \caption{Binary}
        \label{fig:binary_depth}
    \end{subfigure}
    \begin{subfigure}[t]{0.25\linewidth}
        \centering
        \includegraphics[width=\linewidth]{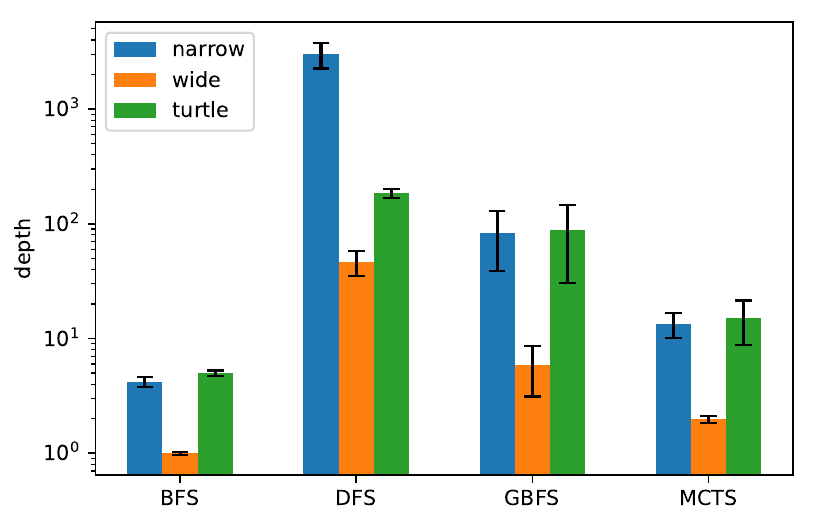}
        \caption{Zelda}
        \label{fig:zelda_depth}
    \end{subfigure}
    \begin{subfigure}[t]{0.25\linewidth}
        \centering
        \includegraphics[width=\linewidth]{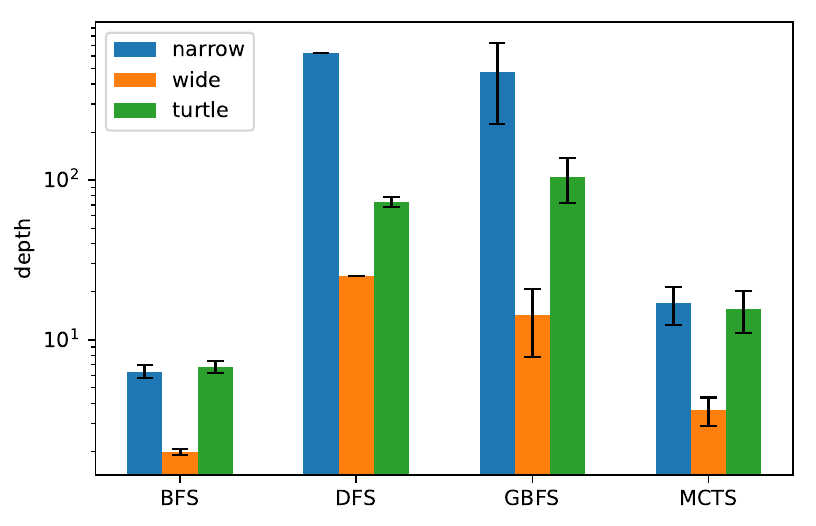}
        \caption{Sokoban}
        \label{fig:sokoban_depth}
    \end{subfigure}
    \caption{The average maximum depth for all the tree search algorithms over 2500 runs. The actual bars are the average value while the black vertical lines show the standard deviation.}
    \label{fig:depth}
\end{figure*}

\subsection{Representations}\label{sec:representation}
Problem representation is critical in search-based PCG~\cite{ashlock2018exploring} as it impacts the speed of generation and the style of the generated content. For example, one way of representing a level generation problem is with a 1 dimensional integer array. This representation lends itself directly to the optimization algorithms where the algorithm is able to modify the level by directly changing a single index in that array. Tree search algorithms need to model this representation within a graph of nodes and connections. We can imagine every map as a node in the space of all possible maps in an entire game, and the connection between these nodes are based on the changes needed to reach that map from the other map. We can then limit these connections to mimic the behavior of optimization algorithms: only one tile is modified to move from node to node. Thus, we can replicate problem representation regardless of the search method used.

In this section, we introduce three different representations (\emph{Narrow}, \emph{Turtle}, and \emph{Wide}) that can transform the generation process into a graph which can be easily traversed using Tree Search Algorithm. Before applying a tree search algorithm, the root node has to be selected. In this work, we select that node randomly, similar to the random starting maps of the optimization algorithms.

\textbf{Narrow Representation} is defined as changing one specified tile at a time. In this approach, the tiles that the algorithm can modify are randomly ordered, and the algorithm can only modify the map in that particular order. This means that each tree node represents the current map and the current modified tile, while the branches represents the modifications that can be done to that tile. This modification decreases the branching factor to be $n$ actions (where $n$ is the number of different game tiles) but it increased the state space by adding the current modified tile as part of the state representation.

\textbf{Turtle Representation} draws parallels to the Turtle Graphics module in the Logo programming language. In this representation, algorithms are given a random initial position within the map. They are allowed to either change its position by moving to any of the neighboring tiles in the four cardinal directions (unless a direction would take them ``out-of-bounds'') or modify this tile to another tile value, and the process repeats. This means that a node in the tree represents the map and the turtle's current position, while the branches are the directional movement and the map modification decisions. Similar to the narrow representation, the action space decreased to be $4+n$ actions (4 cardinal movements and $n$ tile modification actions) but it increased the state space by adding the current modified tile as part of the state representation. 

\textbf{Wide Representation} is inspired from the optimization algorithms' representation. In this representation, the algorithm itself can decide exactly which tiles to modify in any order. For tree search, this means that a node only represents the current map while the branches are equal to the map size multiply by number of possible tiles where it identifies which tile location can be changed and what is the new tile type.

\subsection{Problems}\label{sec:problems}
The problems that we are using for the comparison are taken from the PCGRL framework\footnote{https://github.com/amidos2006/gym-pcgrl}. The framework supports six different problems. In this work, we will focus only on three problems, the same ones used in~\cite{khalifa2020pcgrl}. Each problem is represented as 2D array of tiles.
The heuristic/fitness function is provided through the PCGRL framework for each problem. To evaluate any problem state, the PCGRL framework compares the current state with a reference state (the root state in case of tree search algorithms and a random initial chromosome in case of optimization algorithms). 
For example, if the goal is to create 2D maze with a long path. If the initial state has a path length of 5 and the current state has a path length of 10, then the fitness/heuristic will be 5. On the other hand, if the current state has a path length of 2, the fitness/heuristic will be -3. If there are multiple goals for the problem, the heuristic/fitness from each goal is combined as a weighted sum\footnote{check the repository for detailed implementation of the fitness/heuristic functions: https://github.com/amidos2006/tsxoa}.

\textbf{Binary} is the simplest problem: the goal is to create a binary map layout where all the empty tiles are connected and the longest shortest path between any two points in the map increases by at least $X$ ($X$ = 20 in this work). 

\textbf{Zelda} is inspired by the GVGAI~\cite{perez2016general} version of the dungeon system in The Legend of Zelda (Nintendo, 1986). The player has to collect a key and the reach the door without dying from the moving enemies. The generator must take into consideration the goal of the game and must try to generate a playable level. A playable level must have 1 player, 1 key, 1 door, all tiles fully connected, enemy should be $Y$ step away from player
and the path between the player and the key as well as the key and the door must be at least $X$ steps ($Y$ = 5,$X$ = 20 in this work). 

\textbf{Sokoban} is a port of the famous Japanese game by the same name (Thinking Rabbit, 1982). The goal of the game is to push every crate to a goal location. To achieve that goal, the generated levels has to have 1 player, number of crates equal to number of targets, and can be solved using A* algorithm with at least $X$ steps ($X$ = 20 in this work). 

\begin{figure*}
    \centering
    \begin{subfigure}[t]{0.25\linewidth}
        \centering
        \includegraphics[width=\linewidth]{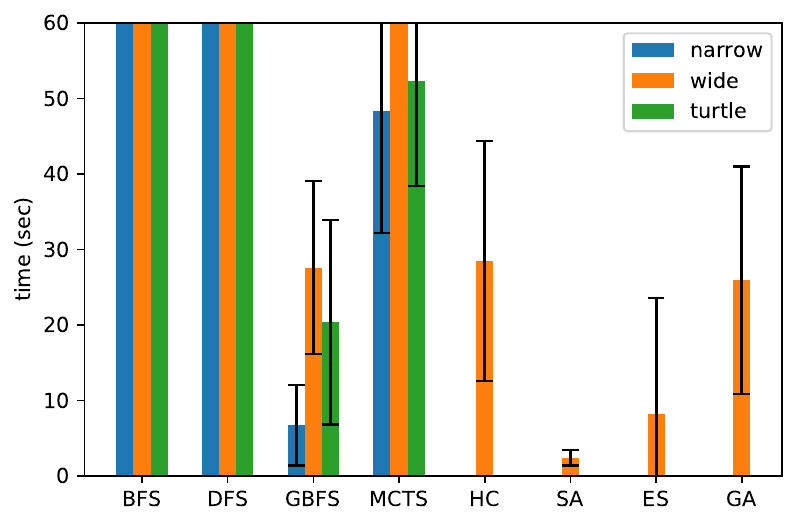}
        \caption{Binary}
        \label{fig:binary_time}
    \end{subfigure}
    \begin{subfigure}[t]{0.25\linewidth}
        \centering
        \includegraphics[width=\linewidth]{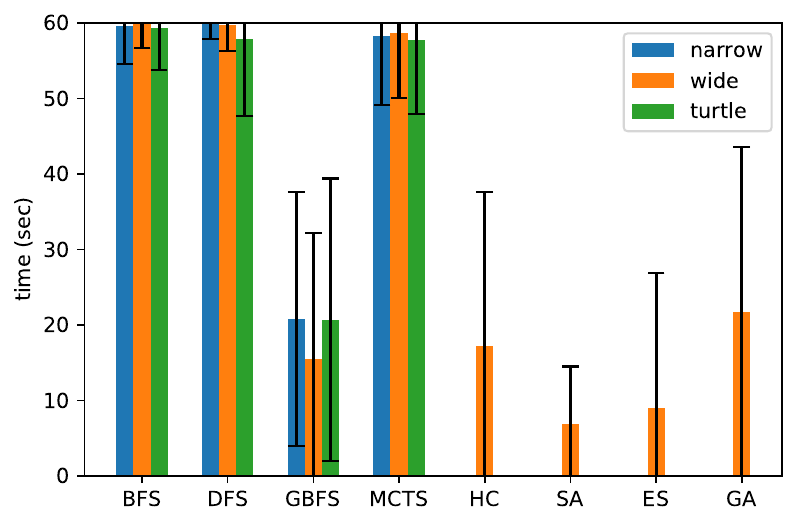}
        \caption{Zelda}
        \label{fig:zelda_time}
    \end{subfigure}
    \begin{subfigure}[t]{0.25\linewidth}
        \centering
        \includegraphics[width=\linewidth]{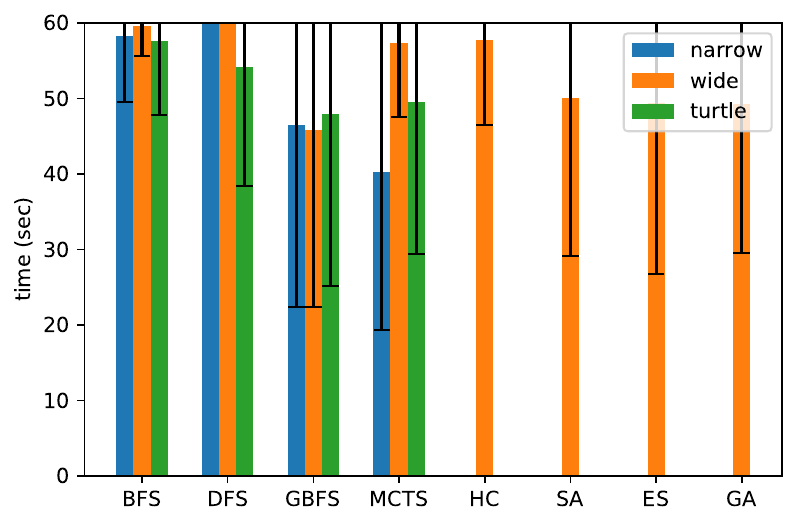}
        \caption{Sokoban}
        \label{fig:sokoban_time}
    \end{subfigure}
    \caption{The average time in seconds that each algorithm takes to run over all 2500 runs. The actual bars are the average value while the black vertical lines show the standard deviation.}
    \label{fig:time}
\end{figure*}

\section{Experiments}\label{sec:experiments}


Our generators are configured to create maps of size 14x14 for Binary, 11x7 for Zelda and 5x5 for Sokoban excluding boundaries.
We run all the eight algorithms on all the three problems using different representations. Optimization algorithm are only run using wide representation as it is the direct representation used for optimization algorithms in all the previous work. 
We introduce the narrow and turtle representations to assist Tree Search algorithm manage the large branching factor in the graph (392 for binary, 616 for Zelda, and 125 for Sokoban) which optimization algorithm does not have problem with.

We run each experiment for 2500 runs, where each run was capped at 60 seconds. If the algorithm finds the solution before hand, it terminates and returns the solution, otherwise it continues till time out. For MCTS, the $C$ value is set to $5$ to balance with the big values of exploitation term. We also assign the rollout length as 40\% of the size of the problem (78 for Binary, 40 for Zelda, and 10 for Sokoban). This variable length is just to allow the MCTS rollout to have an effect on the map, as the bigger the maps gets the deeper they need to be explored randomly to have an effect on the level output. For SA, we start with temperature equal to $10$ and cooling rate equal to $0.99$. For ES, we use $\mu$ equal to $10$ and $\lambda$ equal to $20$. Finally, GA uses population of size $30$ with elitism of size $1$; new individuals are generated either by crossover (80\% chance) followed by mutation (5\%), or only mutation. Rank selection is used to select parents. For mutation (for all optimization algorithms), we are using single point mutation where a single tile is picked and changed randomly. For crossover (only GA), we are using two point crossover where two points on the map (as a 1D string) are picked and the values between these points are swapped. These values are chosen by earlier experiments that lead to the best result.

\section{Results}\label{sec:results}

Figure~\ref{fig:solutions} shows the performance of all the different algorithms on our 3 different problems using the possible representations. For the tree search algorithms, Greedy algorithms such as GBFS outperformed most of the other algorithms, its performance was the same between the different representations on all the problems. This is likely due to the sensitive heuristic function that we are using for our problems. Any change in the map usually corresponds to a change in the heuristic.

On the other hand, BFS and DFS performed extremely low (almost zero on all the problems and representations) which was expected. BFS is unable to search very deep within the tree where most solutions reside (see Figure~\ref{fig:depth}). BFS depth is very low compared to most of the other algorithms. On the other hand, DFS can search very deep within the tree but still performs badly. This is due to DFS' tendency to trap itself in bad initial branch paths, leading to wasted computational time before exploring other branches that could lead to the solution.  Figure~\ref{fig:depth} shows the depth of the deepest found node by the algorithm during search regardless it found a solution or not. If the algorithm found a solution, it will terminate the search at that node (which will make it the deepest node found); if it did not find a solution, it will continue exploring until $60$ seconds expire.

\begin{figure}
    \centering
    \begin{subfigure}[t]{.45\linewidth}
        \centering
        \includegraphics[width=\linewidth]{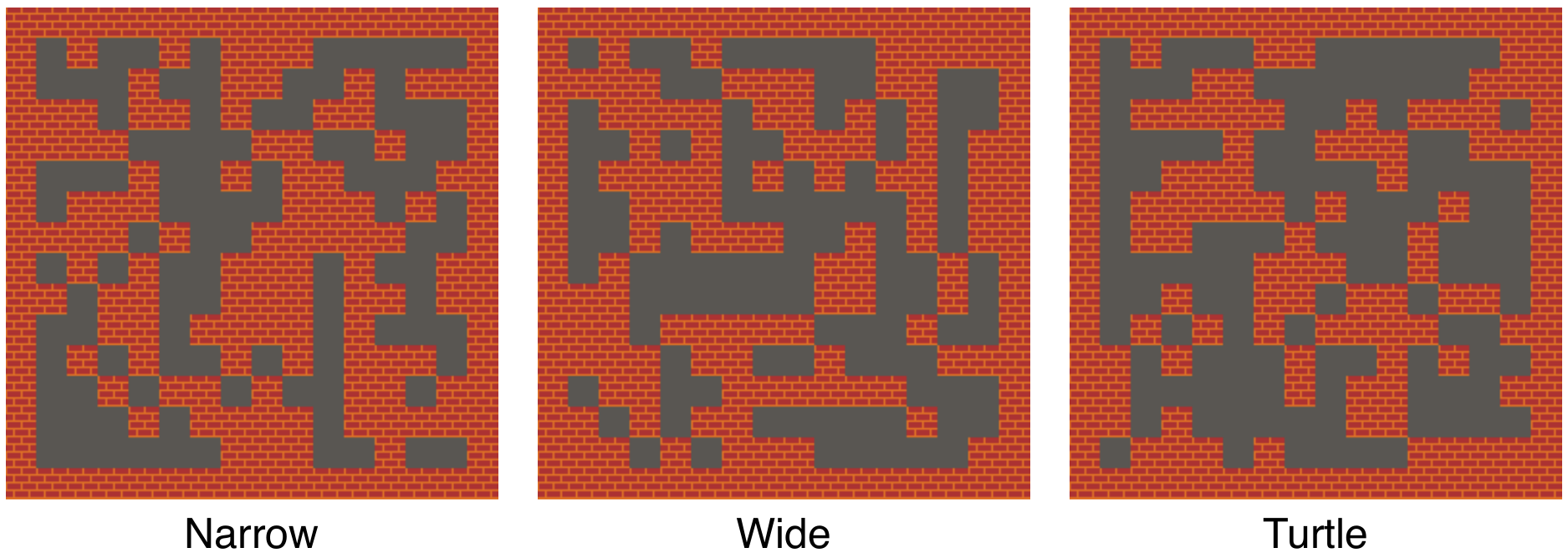}
        \caption{BFS}
        \label{fig:BFS_binary}
    \end{subfigure}
    \begin{subfigure}[t]{.45\linewidth}
        \centering
        \includegraphics[width=\linewidth]{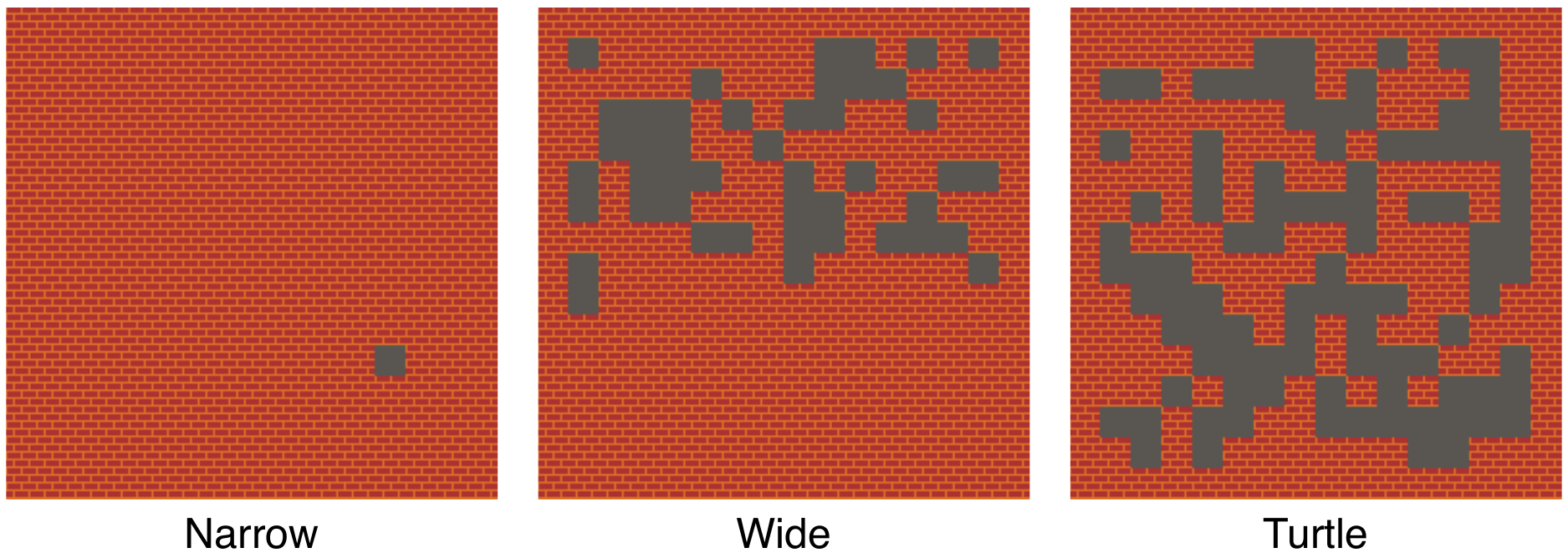}
        \caption{DFS}
        \label{fig:DFS_binary}
    \end{subfigure}
    \begin{subfigure}[t]{.45\linewidth}
        \centering
        \includegraphics[width=\linewidth]{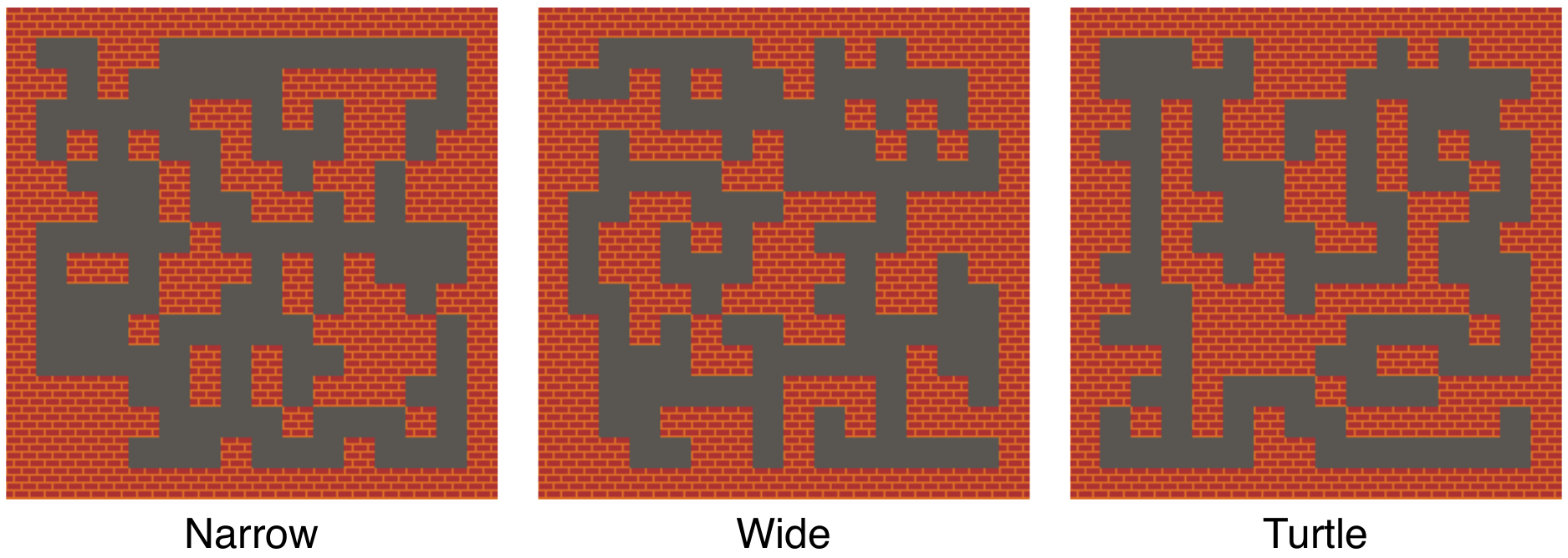}
        \caption{GBFS}
        \label{fig:GBFS_binary}
    \end{subfigure}
    \begin{subfigure}[t]{.45\linewidth}
        \centering
        \includegraphics[width=\linewidth]{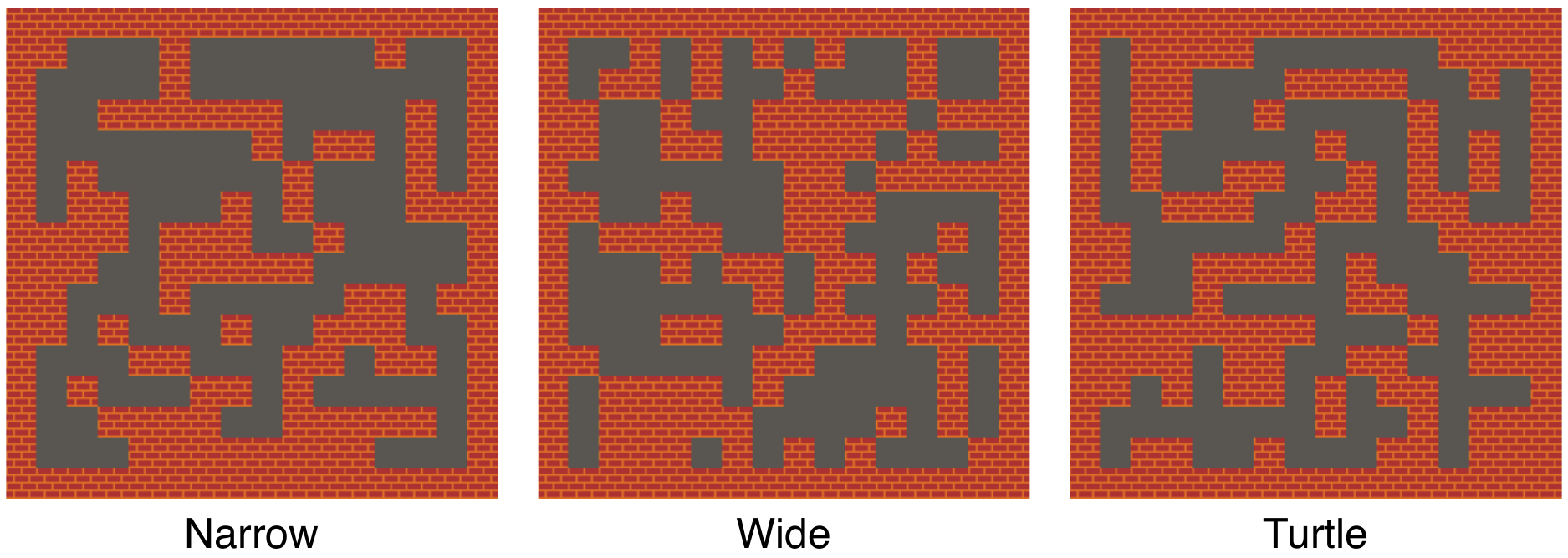}
        \caption{MCTS}
        \label{fig:MCTS_binary}
    \end{subfigure}
    \begin{subfigure}[t]{.45\linewidth}
        \centering
        \includegraphics[width=\linewidth]{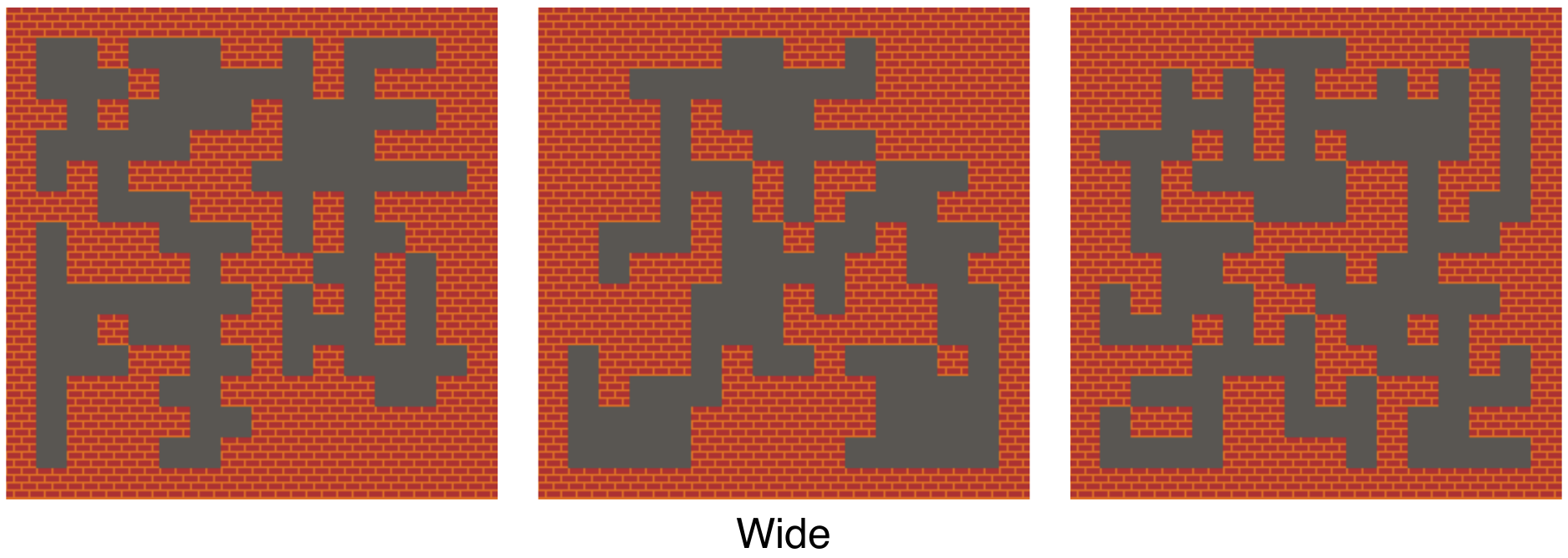}
        \caption{HC}
        \label{fig:HC_binary}
    \end{subfigure}
    \begin{subfigure}[t]{.45\linewidth}
        \centering
        \includegraphics[width=\linewidth]{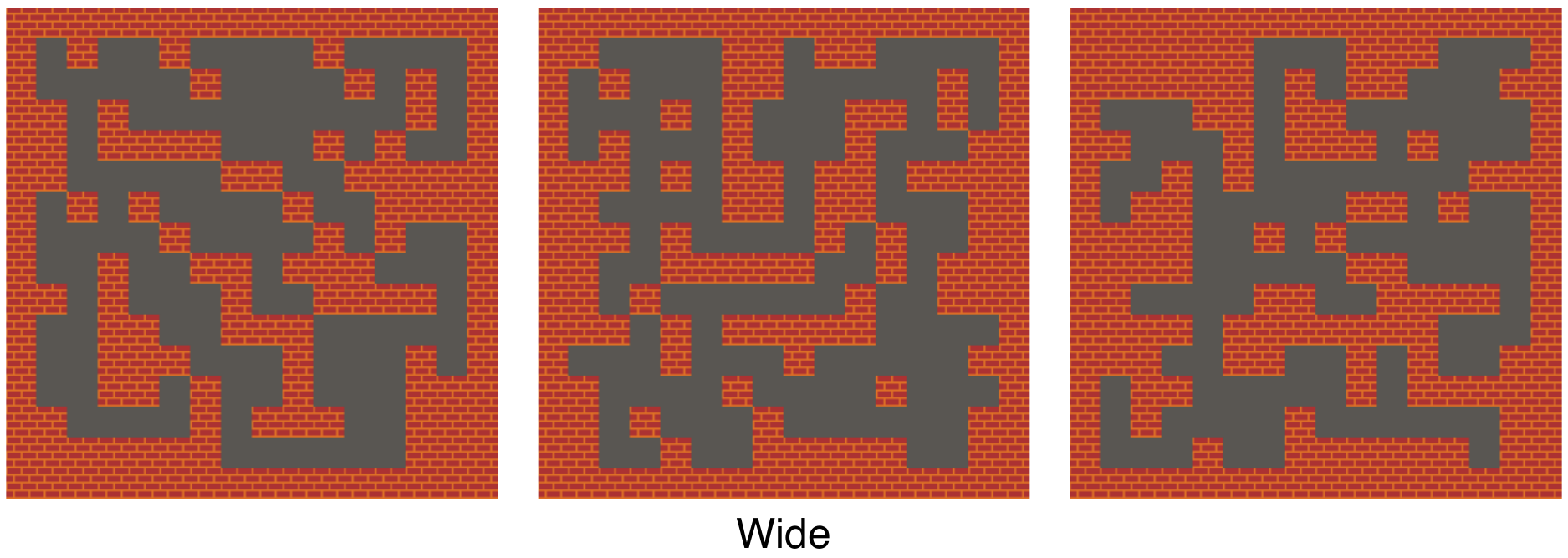}
        \caption{SA}
        \label{fig:SA_binary}
    \end{subfigure}
    \begin{subfigure}[t]{.45\linewidth}
        \centering
        \includegraphics[width=\linewidth]{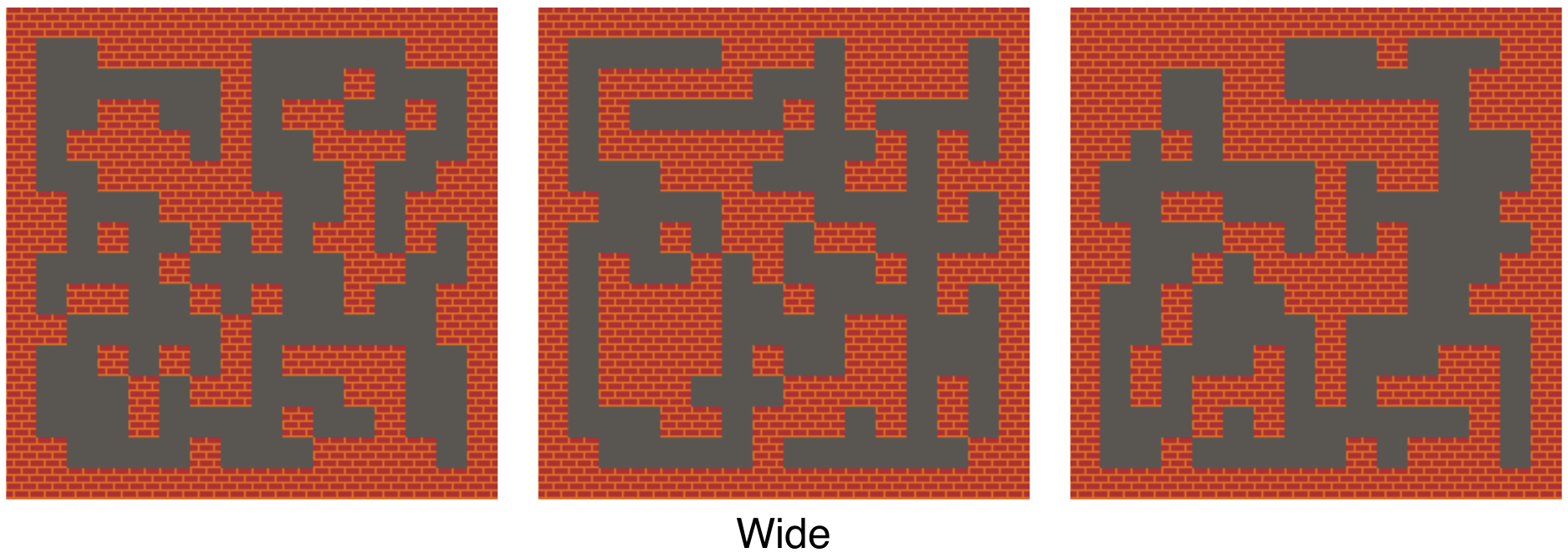}
        \caption{ES}
        \label{fig:ES_binary}
    \end{subfigure}
    \begin{subfigure}[t]{.45\linewidth}
        \centering
        \includegraphics[width=\linewidth]{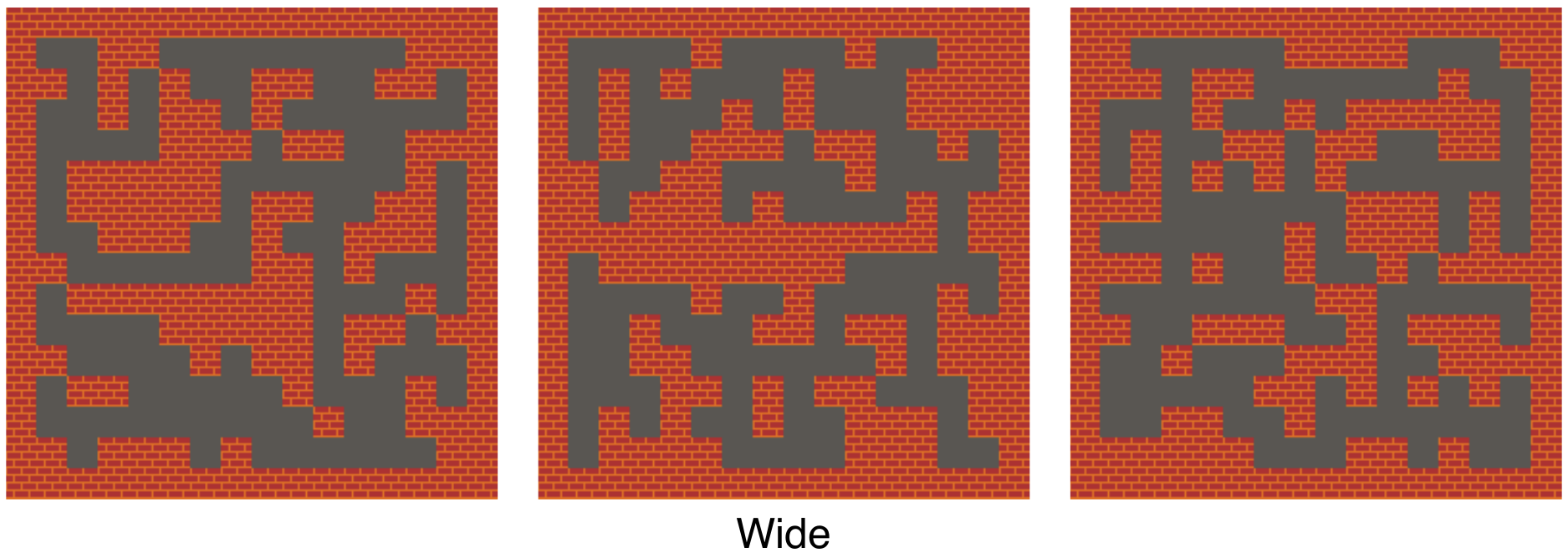}
        \caption{GA}
        \label{fig:GA_binary}
    \end{subfigure}
    \caption{Binary Examples}
    \label{fig:binary_examples}
\end{figure}

MCTS was a different case: it did not search very deep in the tree compared to GBFS due to the exploration factor and the random rollouts. MCTS performed badly on the Binary problem, especially when using wide representation (due to the low depth). On Zelda, MCTS performed worse as well: its performance was almost as bad as DFS and BFS. Looking at the depth, we can see that MCTS built a very shallow tree. We believe this occured due to the big map space (11x7) and the large amount of possible tiles (8 different tiles). The large branching factor lead MCTS to waste most of its computation time on low level nodes as MCTS cannot continue exploring until it expands/simulates/back-propagates all the previous children (which is not the case in GBFS). In Sokoban, MCTS using Narrow representation outperformed all the other algorithms, and at the same time it did not explore deeply in the tree. This is likely due to the deceptive the reward heuristic/fitness landscape of Sokoban~\cite{anderson2018deceptive}. Deceptive landscapes were not a issue in the first two problems (Binary and Zelda) as the heuristic was usually provided accurate guidance toward success. In Sokoban, the heuristic sometimes does not always lead the generator toward playable levels. For example: the number of crates equalling the number of targets and that each of these are reachable from the player's location are a good indicators for playability. However, neither guarantee that the level is beatable. Crates could be reachable but be stuck beside a wall or another crate. GBFS might waste more time on searching a branch that leads to a non playable level compared to MCTS (which estimates how good or bad a branch is before committing to it).
On the other hand, the MCTS turtle representation did not perform well. The random rollouts are likely a leading factor here: they are less effective as 4 of the selected actions are agent movements, leading to less effective sampling and changes in the heuristic estimation of that branch.

The optimization algorithms generally perform similarly on all the problems. We were surprised to see that SA, in spite of being a single point optimization algorithm, performs slightly better than the other optimization algorithms.
We think the temperature is helping it to explore the space faster and reach a good result, while HC, ES, and GA converge pretty fast to a local optimal solution and cannot escape it. We think this can be easily solved by increasing the population size for ES and GA.

\begin{figure}
    \centering
    \begin{subfigure}[t]{.45\linewidth}
        \centering
        \includegraphics[width=\linewidth]{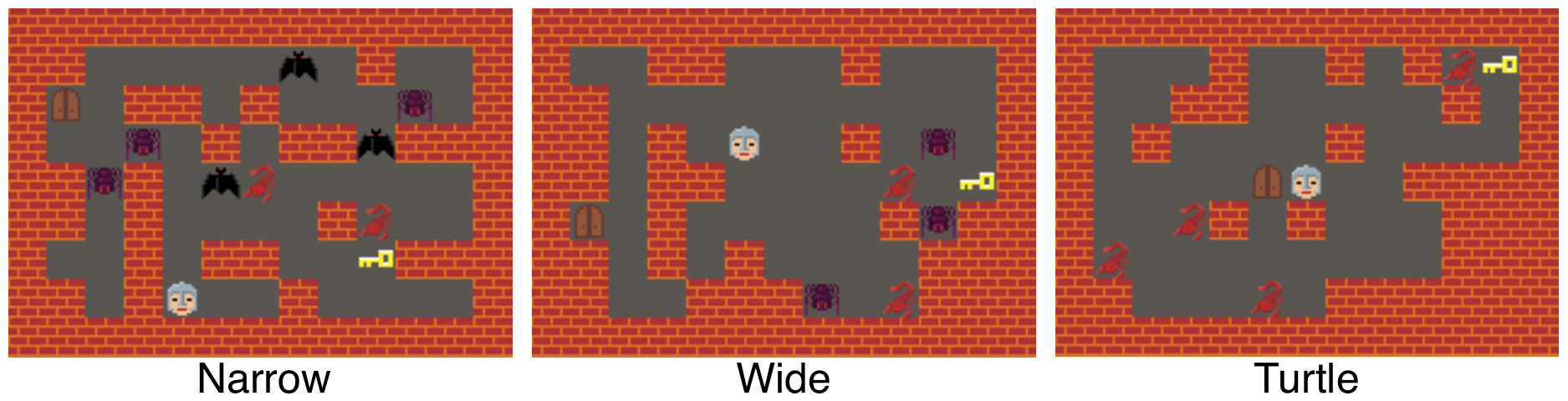}
        \caption{BFS}
        \label{fig:BFS_zelda}
    \end{subfigure}
    \begin{subfigure}[t]{.45\linewidth}
        \centering
        \includegraphics[width=\linewidth]{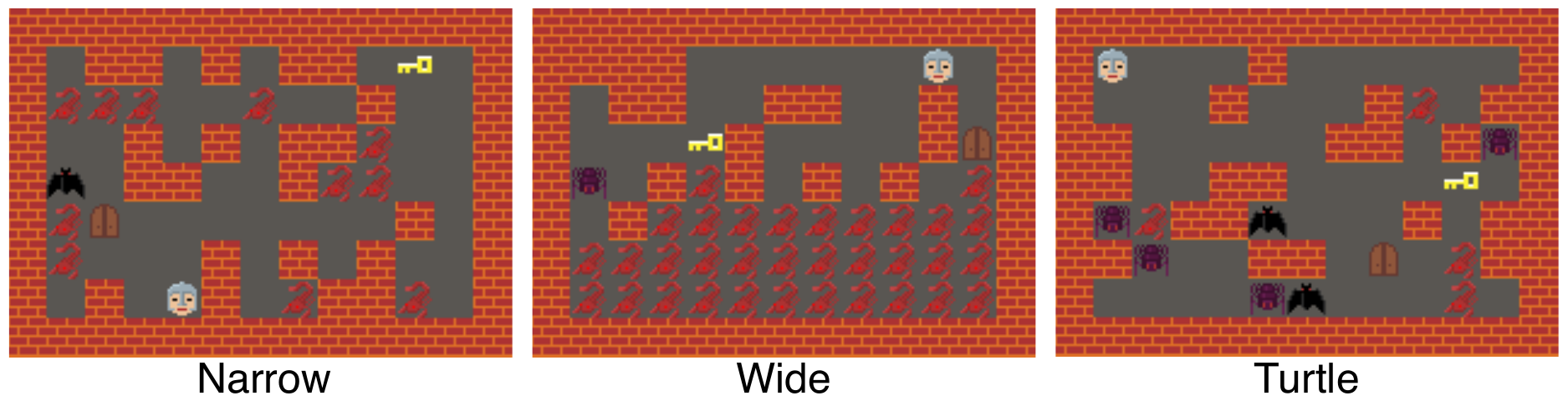}
        \caption{DFS}
        \label{fig:DFS_zelda}
    \end{subfigure}
    \begin{subfigure}[t]{.45\linewidth}
        \centering
        \includegraphics[width=\linewidth]{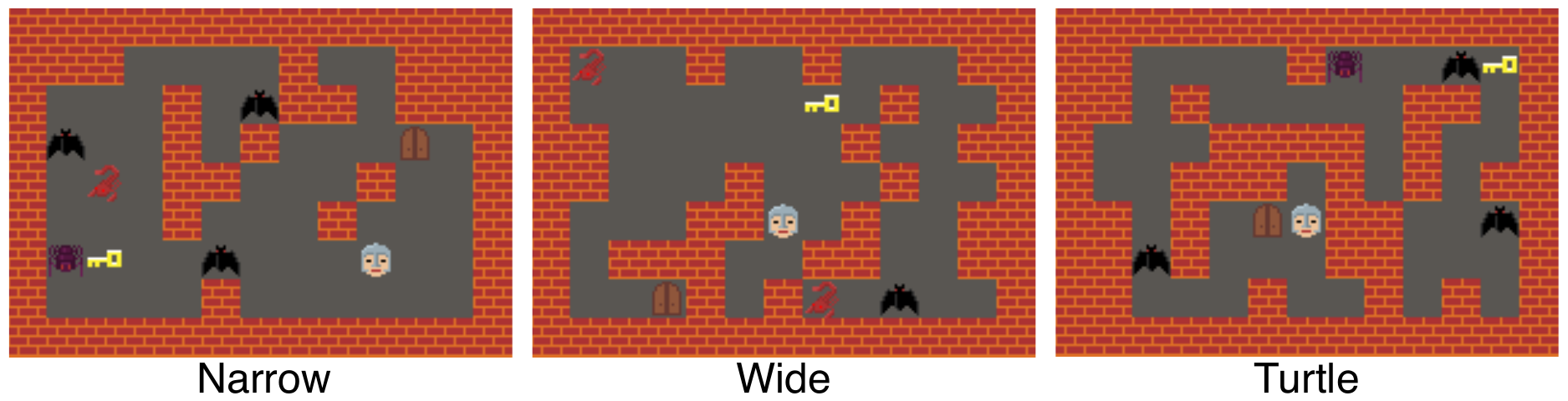}
        \caption{GBFS}
        \label{fig:GBFS_zelda}
    \end{subfigure}
    \begin{subfigure}[t]{.45\linewidth}
        \centering
        \includegraphics[width=\linewidth]{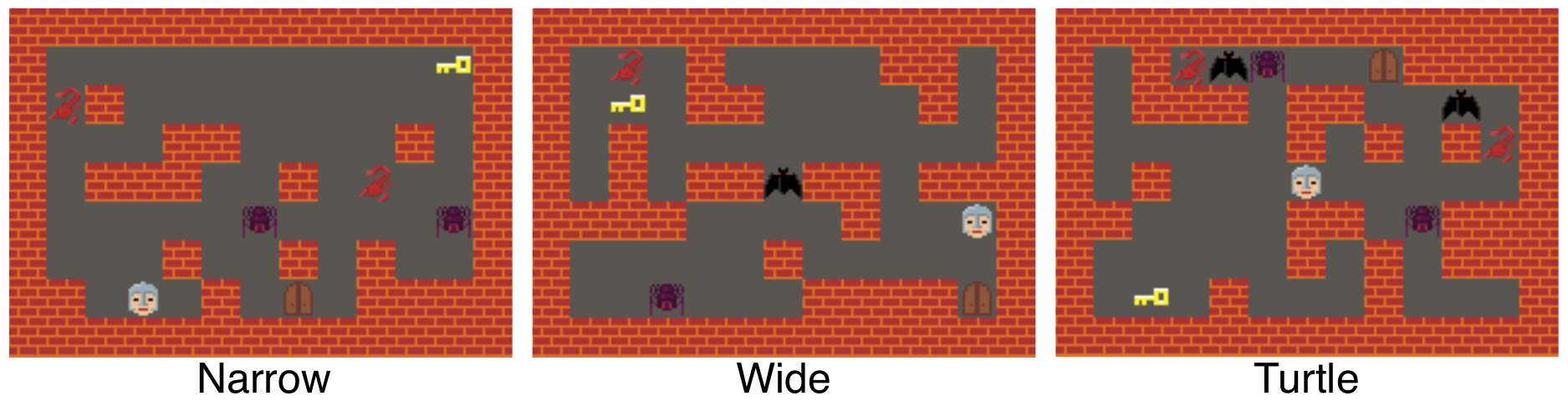}
        \caption{MCTS}
        \label{fig:MCTS_zelda}
    \end{subfigure}
    
    \begin{subfigure}[t]{.45\linewidth}
        \centering
        \includegraphics[width=\linewidth]{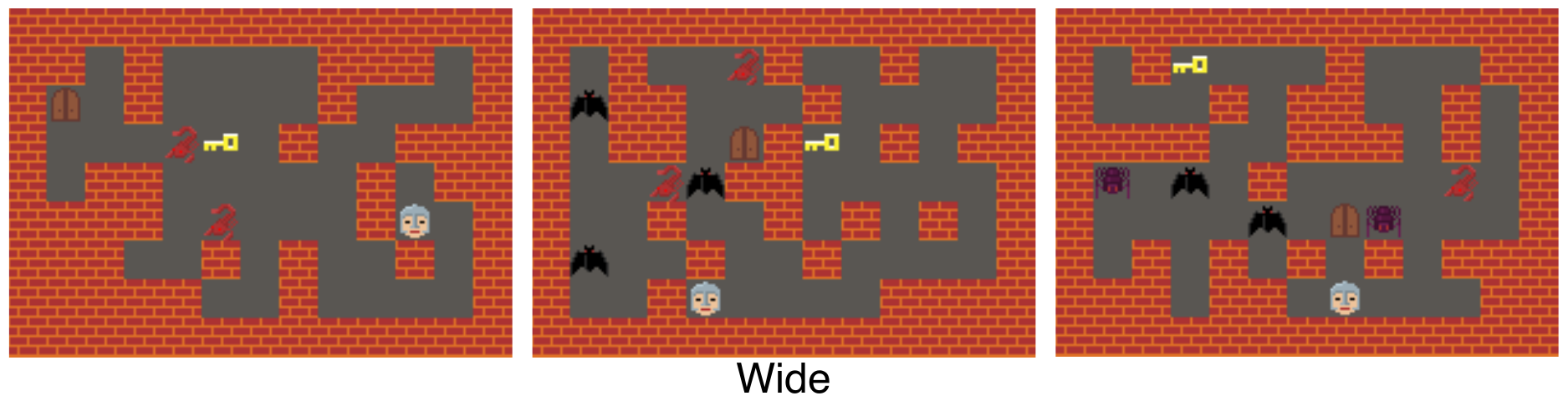}
        \caption{HC}
        \label{fig:HC_zelda}
    \end{subfigure}
    \begin{subfigure}[t]{.45\linewidth}
        \centering
        \includegraphics[width=\linewidth]{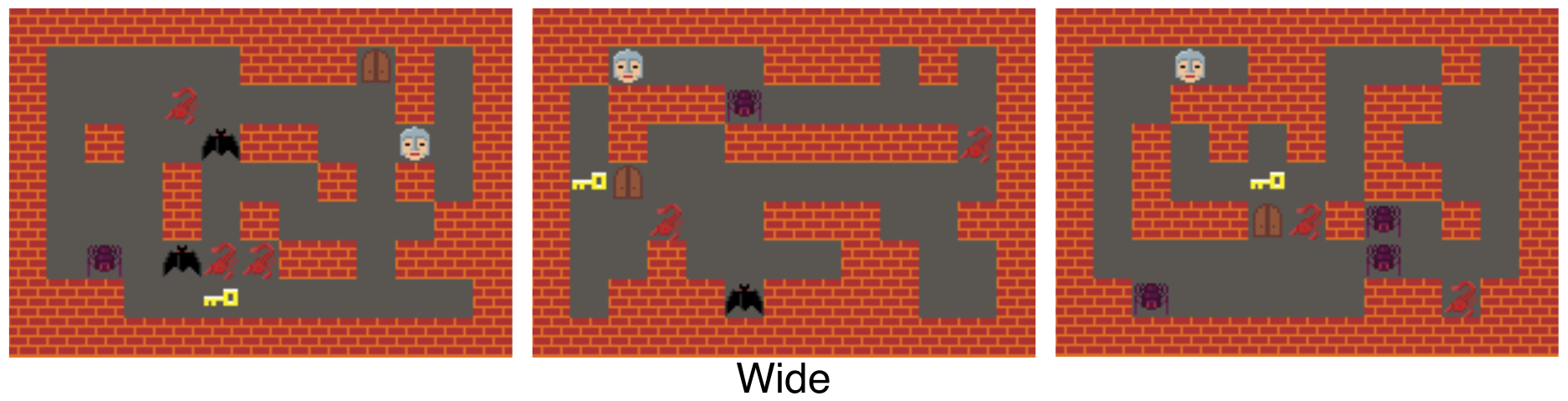}
        \caption{SA}
        \label{fig:SA_zelda}
    \end{subfigure}
    \begin{subfigure}[t]{.45\linewidth}
        \centering
        \includegraphics[width=\linewidth]{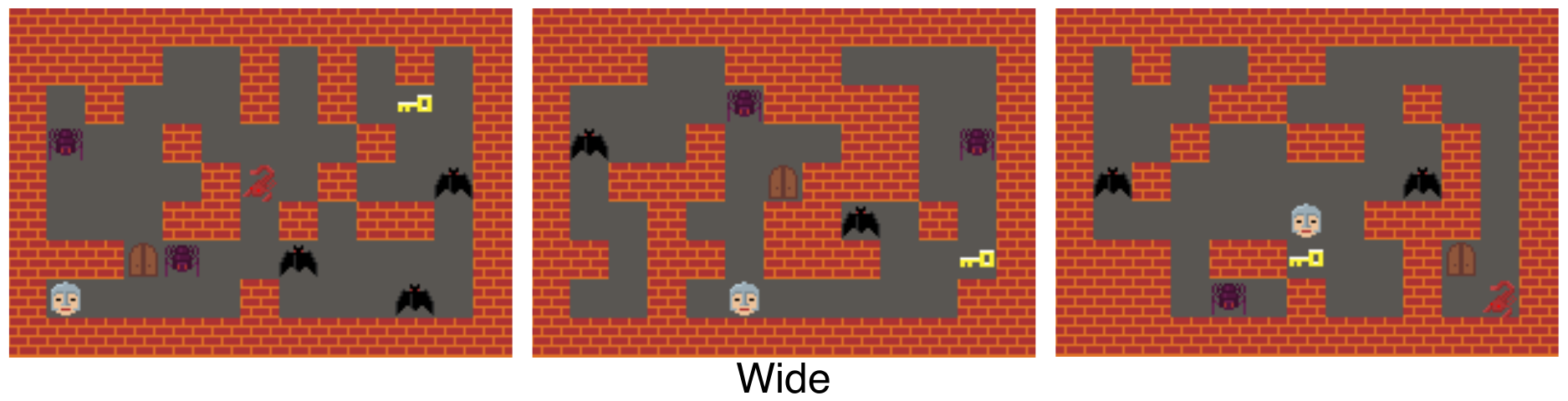}
        \caption{ES}
        \label{fig:ES_zelda}
    \end{subfigure}
    \begin{subfigure}[t]{.45\linewidth}
        \centering
        \includegraphics[width=\linewidth]{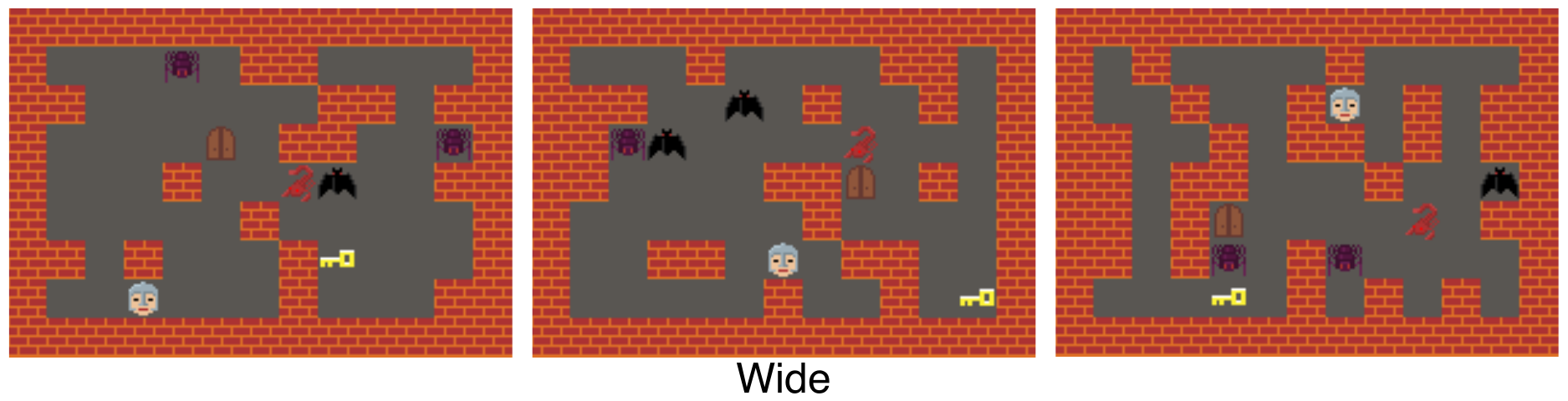}
        \caption{GA}
        \label{fig:GA_zelda}
    \end{subfigure}
    \caption{Zelda Examples}
    \label{fig:zelda_examples}
\end{figure}

An advantage GBFS has over greedy HC algorithm is that GBFS does not get stuck since it can always roll back to any other path to explore if it reaches a local maxima, while HC will always get stuck as it does not keep track of previous explored solutions. This is a classic trade off between memory and performance where GBFS uses more memory which leads better performance than HC algorithm.

Figure~\ref{fig:time} shows the average time in seconds that each algorithm needs to find solution. The optimization algorithms take less time on average to find the results, which was not surprising looking at their performance. GBFS is the only algorithm that can be compared in time due to its greedy nature of visiting the nodes that leads it to the solution quickly (except for Sokoban where it easily gets deceived and does not find the solution). BFS, DFS and wide MCTS are always near the cut off time (60 seconds): they are never able to find a solution before timeout.

\begin{figure}
    \centering
    \begin{subfigure}[t]{.45\linewidth}
        \centering
        \includegraphics[width=\linewidth]{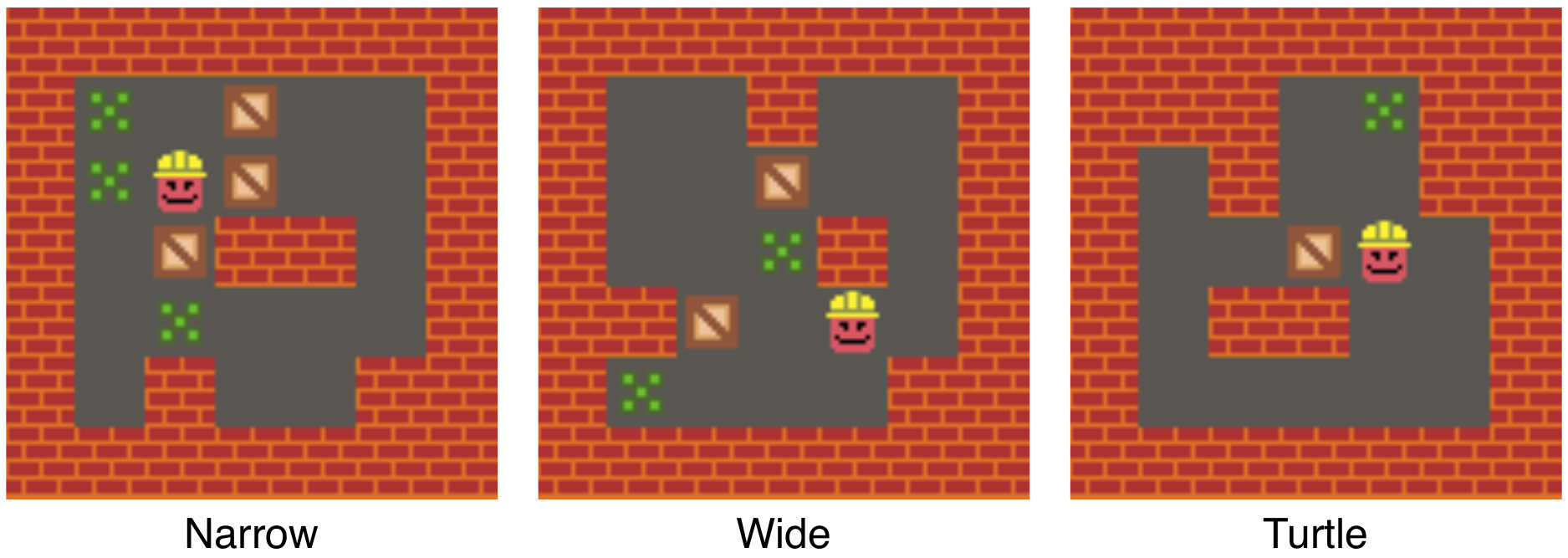}
        \caption{BFS}
        \label{fig:BFS_sokoban}
    \end{subfigure}
    \begin{subfigure}[t]{.45\linewidth}
        \centering
        \includegraphics[width=\linewidth]{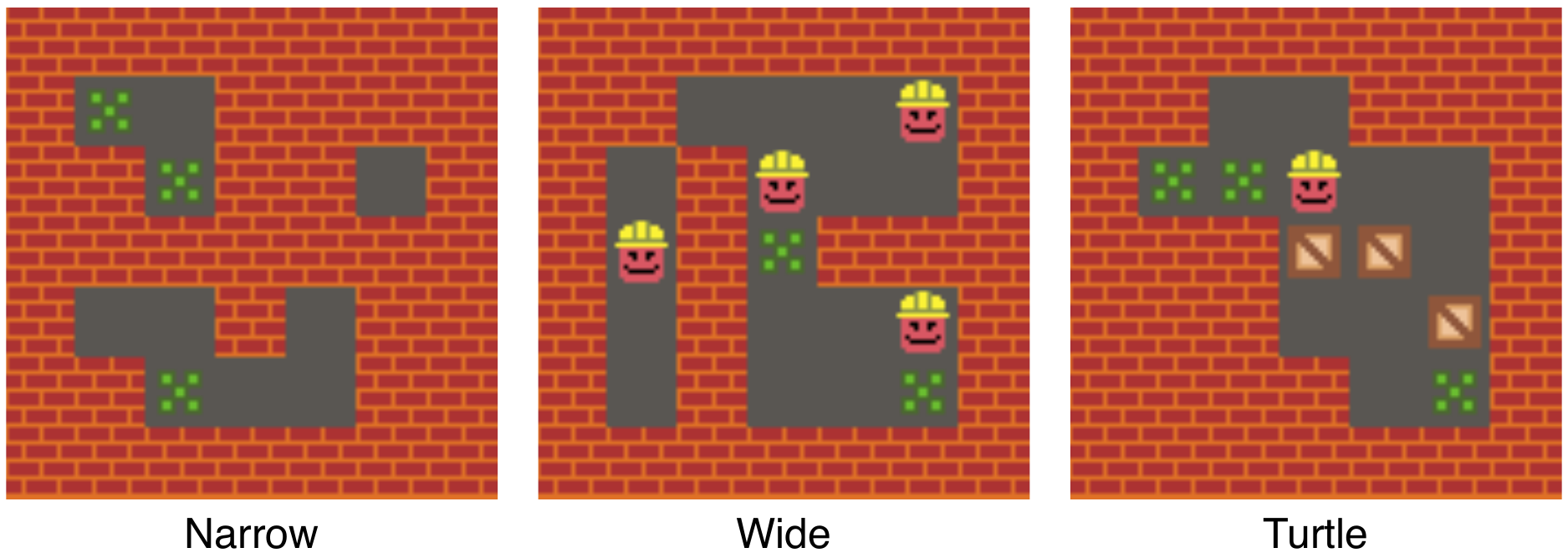}
        \caption{DFS}
        \label{fig:DFS_sokoban}
    \end{subfigure}
    \begin{subfigure}[t]{.45\linewidth}
        \centering
        \includegraphics[width=\linewidth]{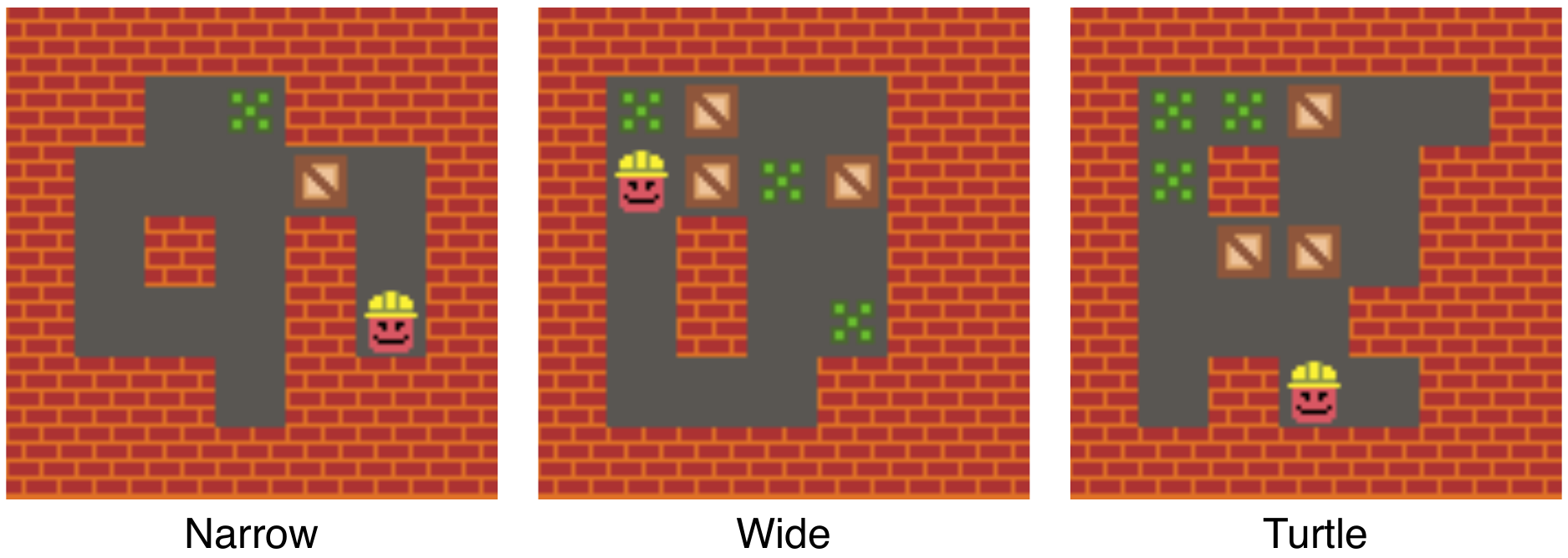}
        \caption{GBFS}
        \label{fig:GBFS_sokoban}
    \end{subfigure}
    \begin{subfigure}[t]{.45\linewidth}
        \centering
        \includegraphics[width=\linewidth]{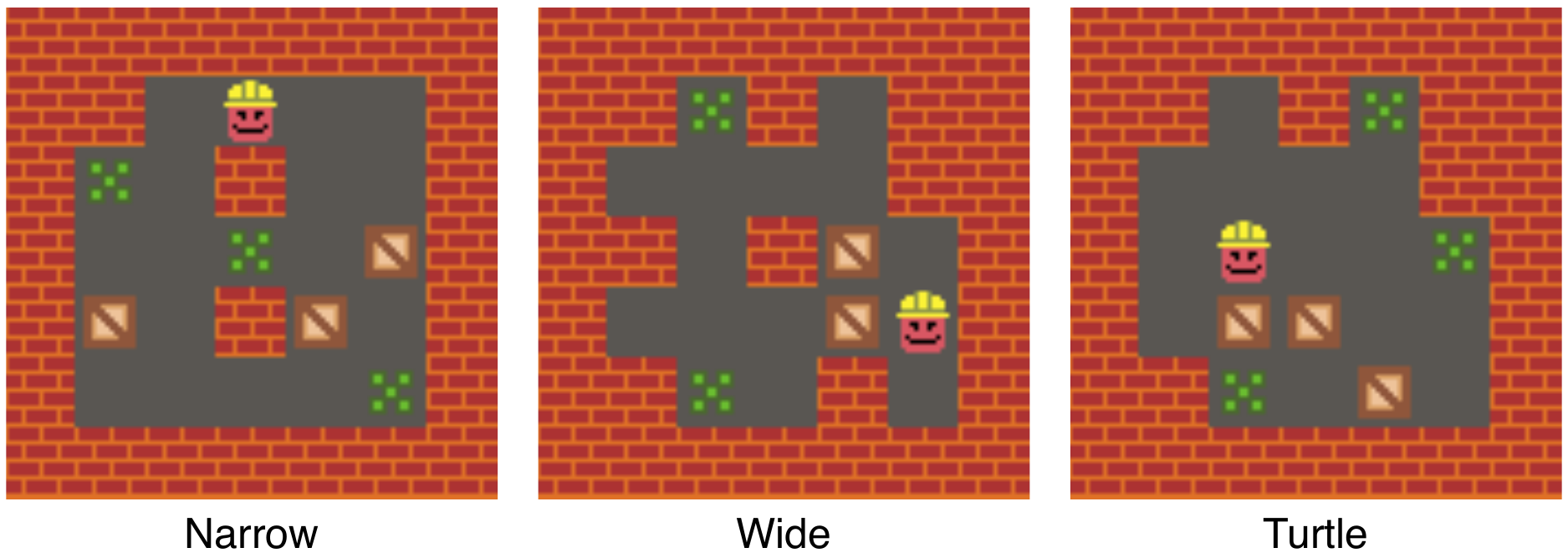}
        \caption{MCTS}
        \label{fig:MCTS_sokoban}
    \end{subfigure}
    \begin{subfigure}[t]{.45\linewidth}
        \centering
        \includegraphics[width=\linewidth]{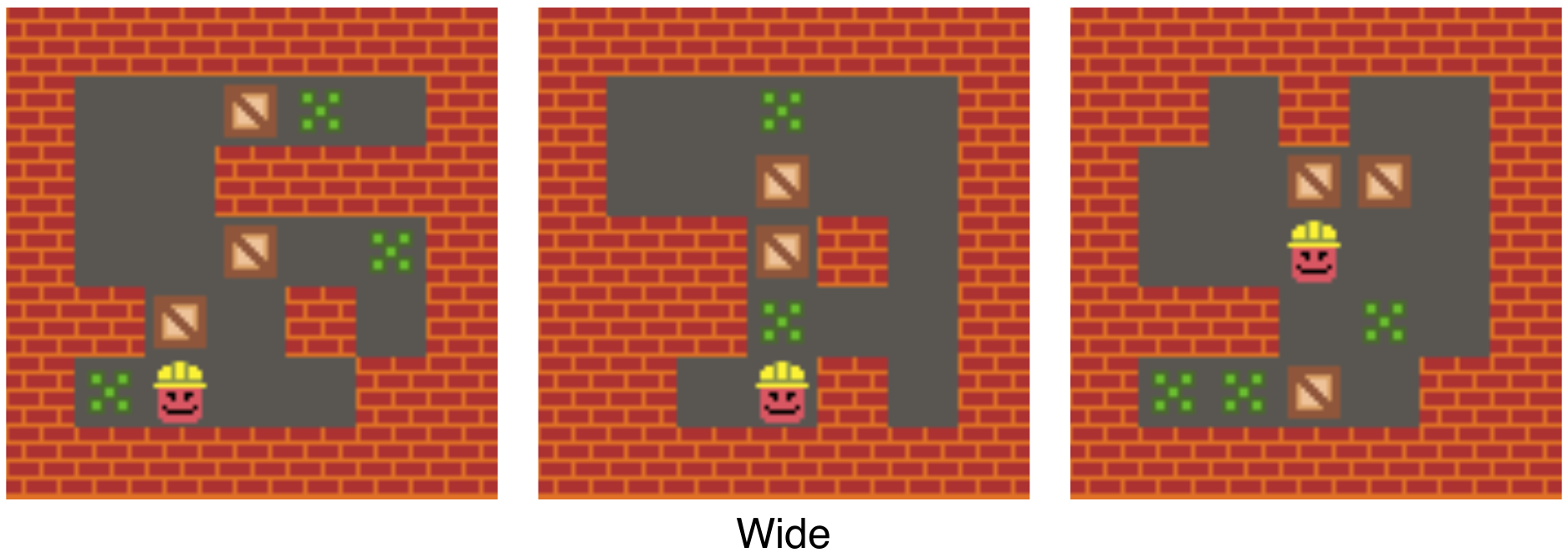}
        \caption{HC}
        \label{fig:HC_sokoban}
    \end{subfigure}
    \begin{subfigure}[t]{.45\linewidth}
        \centering
        \includegraphics[width=\linewidth]{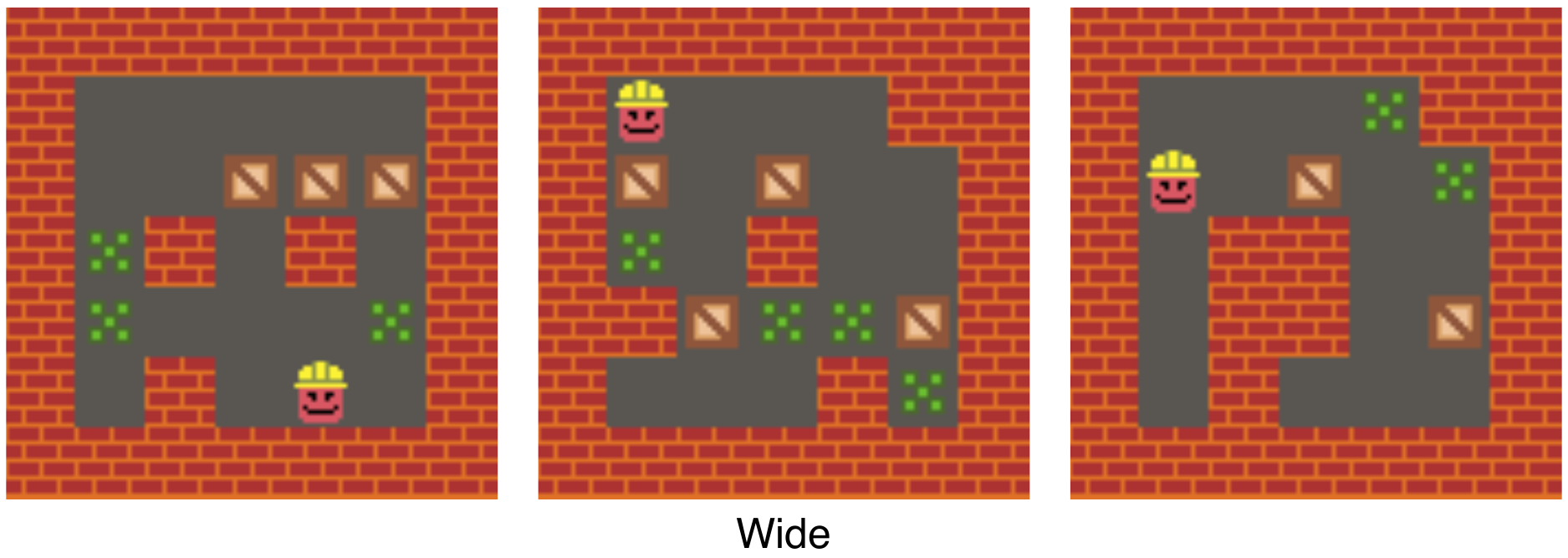}
        \caption{SA}
        \label{fig:SA_sokoban}
    \end{subfigure}
    \begin{subfigure}[t]{.45\linewidth}
        \centering
        \includegraphics[width=\linewidth]{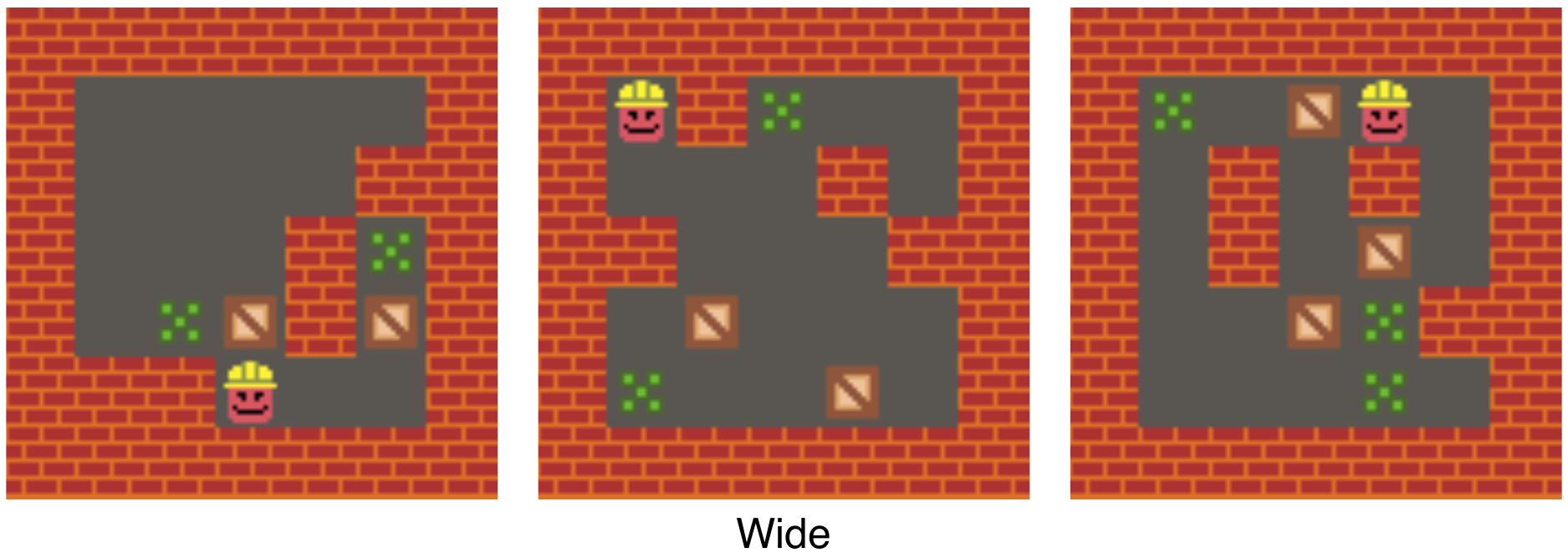}
        \caption{ES}
        \label{fig:ES_sokoban}
    \end{subfigure}
    \begin{subfigure}[t]{.45\linewidth}
        \centering
        \includegraphics[width=\linewidth]{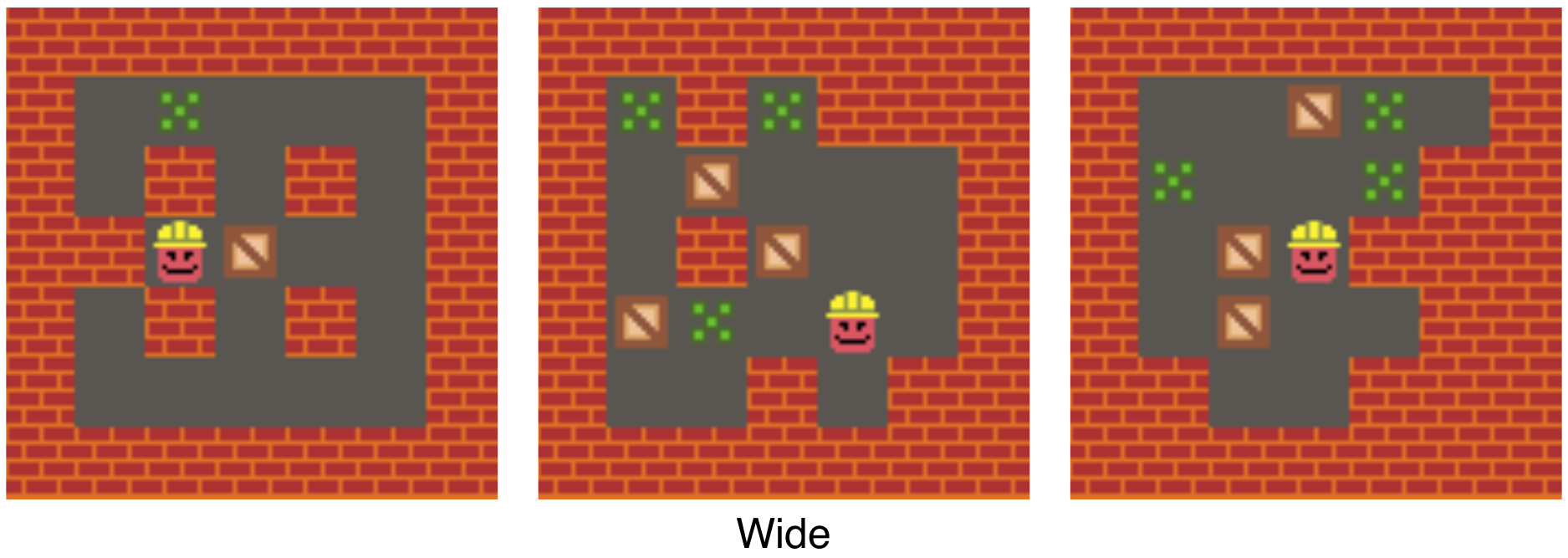}
        \caption{GA}
        \label{fig:GA_sokoban}
    \end{subfigure}
    \caption{Sokoban Examples}
    \label{fig:sokoban_examples}
\end{figure}

Figure~\ref{fig:binary_examples}, \ref{fig:zelda_examples}, and \ref{fig:sokoban_examples} shows examples of the best found levels for every different algorithm on all the three problems. BFS and DFS did not find any solution in most of the problems but it is interesting to look at their best found level. Since the DFS algorithm sticks with a certain action until the end of the tree it forces it to repeat a certain element a lot. 

In Figure~\ref{fig:binary_examples}, The BFS maps looks like almost finished very few tiles are not connected, while the DFS algorithm best solutions cover the map with solid tiles such that it ends with having a single connected region which is interesting. Similar behavior can be seen in Zelda (figure~\ref{fig:zelda_examples}) and Sokoban (figure~\ref{fig:sokoban_examples}) regarding the unfinished DFS levels. Another notable change is SA solutions, they are usually have a different structural aesthetics compared to the rest. It can be easily seen in Zelda, the levels looks more horizontally connected than other levels. We theorize that temperature variable allows SA to reach areas that is not easily reached by being greedy or take longer time during hierarchical exploration.

\section{Conclusion}\label{sec:conclusion}
In this paper, we compared four tree search algorithms to four optimization algorithms in level generation for three different problems (Binary, Zelda, and Sokoban). We introduced two new representations (Narrow and Turtle) to battle the problem of the high branching factor that can affect tree search performance badly. We found that GBFS performed very similar to optimization algorithms. While MCTS did not perform well on Binary and Zelda problem, it exceeded our expectations on the Sokoban problem. We think that the random sampling capabilities allows MCTS to avoid getting stuck in a local optima. 

In most cases, it is hard to notice the difference between the generated levels. However, there are subtle differences, suggesting that each algorithm has its own ``style'' as an effect of the way the algorithm searches for a solution.
While the optimization algorithms typically reach quantitatively better results, the right tree search algorithms paired with the right representation can solve the same content generation problem in a different way. 

We would like to take this work further and investigate population-based tree search algorithms, treating the domain as a graph with multiple starting points similar to the work by \citeauthor{browne2013uct}~\shortcite{browne2013uct}. We would also like to extend this work by using bigger maps and more complex problems (such as Super Mario Bros level generation) and compare new state of the art algorithms in both tree search and optimization algorithms. Another direction is to test these techniques on different domains and generative problems and see how well they translate (narrative generation, character generation, sprite generation, rule generation, etc). 

\section*{Acknowledgements}
Ahmed Khalifa acknowledges the financial support from NSF grant (Award number 1717324 - ``RI: Small: General Intelligence through Algorithm Invention and Selection.''). Michael Cerny Green acknowledges the financial support of the SOE Fellowship from NYU Tandon School of Engineering.
\newpage
\bibliography{aaaibiblo}
\bibliographystyle{aaai}

\end{document}